\documentclass[conference]{IEEEtran}
\IEEEoverridecommandlockouts
\usepackage{cite}
\usepackage{amsmath,amssymb,amsfonts}
\usepackage{algorithmic}
\usepackage{graphicx}
\usepackage{textcomp}
\usepackage{xcolor}

\usepackage{float}
\usepackage{caption}
\usepackage{subcaption}
\usepackage{multirow}
\usepackage{stfloats}

\usepackage[normalem]{ulem}
\useunder{\uline}{\ul}{}

\def\BibTeX{{\rm B\kern-.05em{\sc i\kern-.025em b}\kern-.08em
    T\kern-.1667em\lower.7ex\hbox{E}\kern-.125emX}}
\begin{document}

\title{Attenuation-Aware Weighted Optical Flow with Medium Transmission Map for Learning-based Visual Odometry in Underwater terrain\\
}

\author{
\IEEEauthorblockN{1\textsuperscript{st} Nguyen Gia Bach}
\IEEEauthorblockA{\textit{Graduate School of} \\
\textit{Engineering and Science} \\
\textit{Shibaura Institute of}\\
\textit{Technology, Japan}\\
nb23505@shibaura-it.ac.jp}
\and
\IEEEauthorblockN{2\textsuperscript{nd} Chanh Minh Tran}
\IEEEauthorblockA{\textit{College of Engineering} \\
\textit{Shibaura Institute of}\\
\textit{Technology, Japan}\\
tran.chanh.r4@sic.shibaura-it.ac.jp}
\and
\IEEEauthorblockN{3\textsuperscript{rd} Eiji Kamioka}
\IEEEauthorblockA{\textit{Graduate School of} \\
\textit{Engineering and Science} \\
\textit{Shibaura Institute of}\\
\textit{Technology, Japan}\\
kamioka@shibaura-it.ac.jp}
\and
\IEEEauthorblockN{4\textsuperscript{th} Phan Xuan Tan}
\IEEEauthorblockA{\textit{Graduate School of} \\
\textit{Engineering and Science} \\
\textit{Shibaura Institute of}\\
\textit{Technology, Japan}\\
tanpx@shibaura-it.ac.jp}
}

\maketitle

\begin{abstract}
This paper addresses the challenge of improving learning-based monocular visual odometry (VO) in underwater environments by integrating principles of underwater optical imaging to manipulate optical flow estimation. Leveraging the inherent properties of underwater imaging, the novel wflow-TartanVO is introduced, enhancing the accuracy of VO systems for autonomous underwater vehicles (AUVs). The proposed method utilizes a normalized medium transmission map as a weight map to adjust the estimated optical flow for emphasizing regions with lower degradation and suppressing uncertain regions affected by underwater light scattering and absorption. wflow-TartanVO does not require fine-tuning of pre-trained VO models, thus promoting its adaptability to different environments and camera models. Evaluation of different real-world underwater datasets demonstrates the outperformance of wflow-TartanVO over baseline VO methods, as evidenced by the considerably reduced Absolute Trajectory Error (ATE). The implementation code is available at: https://github.com/bachzz/wflow-TartanVO

\end{abstract}


\section{Introduction} \label{section-1}
The ability of an autonomous underwater vehicle (AUV) to know its position and surroundings is of great importance for underwater missions. Vision-based localization and mapping (visual SLAM) is often favored due to advantages such as low-cost and low-power consumption, over other sensors (i.e., IMU, sonars, laser scanners). Visual SLAM involves visual odometry (VO) and global corrections. VO estimates the 6DoF motion of the vehicle based on current and previous visual information. Due to the only access of the local states, the estimation of VO usually leads to drift in localization. Monocular VO using a single camera is the most challenging, causing further drifts due to scale ambiguity issues. In this manner, an accurate VO can reduce the localization drift and minimize the computational need for further global corrections.

Research on monocular VO has been transitioning from geometry-based \cite{b3, b4} to learning-based \cite{b6} methods, aiming to obtain robustness when working on a variety of environments or camera types, and solve the scale-drift issue in the prior approach. Learning-based VO systems tend to rely on a dense optical flow map, expressing dense 2D matching between two frames. Nevertheless, most of the optical flow estimation algorithms fail in degraded weather situations like foggy weather, as assumed and realized in \cite{b10}. In our paper, this assumption is taken further to underwater environment based on that the same approximate image formation model can be applied to both underwater and foggy environments \cite{b11}. The hazing effect in a foggy environment hinders existing optical flow methods from tracking pixel intensities across frames. This hazing effect is caused by the light scattering factor \cite{b13}, whose component can be found in both foggy and underwater image formation models, as shown later in Eq. \ref{eq:1},\ref{eq:3}. 

The estimated optical flow from successive underwater video frames can be unreliable due to degraded regions in each frame, caused by underwater light scattering and absorption. The more degraded regions contain the higher motion uncertainty, and vice versa. Thus, it is intuitive to suppress motions in those uncertain regions, while emphasizing the motions in clearer regions with less degradation factors, to obtain an optical flow map with higher certainty in flow values. The essential question is how much one suppresses or emphasizes a motion vector, which should depend on the hazing degradation level of the pixel corresponding to the motion. The haze is caused by light scattering \cite{b13}, which is partially determined by a medium transmission map in the image formation model. This map indicates the percentage of the scene radiance reaching the camera after being attenuated by light scattering and absorption \cite{b12}. Hence, the haze level of a pixel can be implicitly determined by the medium transmission map value at that pixel. The medium transmission map has been used extensively in the field of underwater image enhancement \cite{b11}\cite{b12}\cite{b13}. In the context of VO, this paper introduces the normalized medium transmission map as a weight map to the estimated optical flow, to obtain an attenuation-aware weighted optical flow map with higher certainty in motion values.

\section{Underwater Optical Imaging Principles} \label{section-2}

Underwater image is affected by the attenuation effect, including multiple degradation factors like light absorption and scattering \cite{b11}. Light traversing through the water is absorbed at different degrees, depending on the wavelength. Red light with a longer wavelength is absorbed the most, and rapidly decreases in intensity when entering further water depth, whereas blue and green lights with shorter wavelengths can go deeper, resulting in a blueish and greenish image. The camera image taken underwater is formed by two main light sources, the directly transmitted light and the background scattered light. The directly transmitted light originates from the target object containing its true light radiance, but will be attenuated by forward-scattering and light absorption, causing color distortion in the image. The forward-scattering effect on pixel intensity is often minor and can be disregarded \cite{b14}. However, the background scattered light does not come from the target object, but is caused by ambient light being scattered by a high number of tiny particles in the water medium, making the output image to be hazy with low contrast.

Underwater physical image formation model can be observed from foggy image model \cite{b11}:

\begin{equation} \label{eq:1}
I(x) = D(x)t(x) + A(1 - t(x))
\end{equation}

in which x denotes pixel coordinates, $I(x)$ is the raw underwater image affected with multiple degradations, $D(x)$ is the true object radiance or can be interpreted as the restored image after removing the underwater degradations, $t(x)$ denotes medium transmission map, and $A$ denotes ambient light. $A(1-t(x))$ indicates the backscattering light, while $D(x)t(x)$ represents the directly transmitted light. $D(x)t(x)$ will be attenuated with a magnitude depending on the attenuation coefficient $\beta$ and the transmission distance $d$: $t(x) = e^{-\beta d}$

Due to wavelength-dependent color absorption in underwater, the underwater image model has been revised to affect color channels differently \cite{b13}:

\begin{equation} \label{eq:3}
I_c(x) = D_c(x)t_c(x) + A_c(1 - t_c(x)), c \in \{R, G, B\}
\end{equation}

As previously explained, the backscattering light $A_c(1-t_c(x))$ exists in both foggy and underwater image models. Since $A_c$ is the ambient light with a constant value across the image, the pixel-dependent transmission map $t_c(x)$ is the main contributor to the backscattering light, thus it can be used to determine the haze level of a pixel in an underwater image.

\section{Methodology} \label{section-4}

\subsection{Overview}

\begin{figure}[t]
\centering
\includegraphics[width=1.\linewidth]{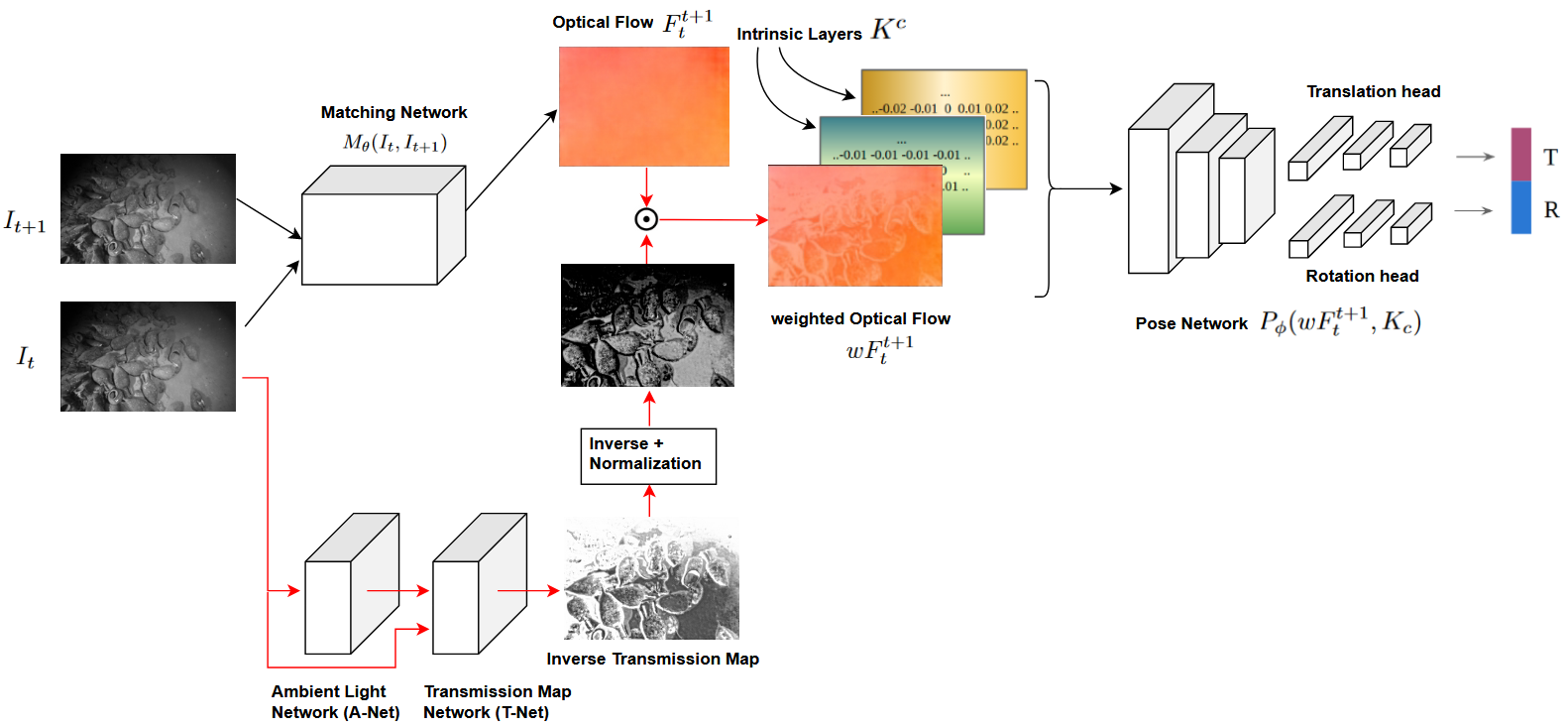}
\caption{Overview of the TartanVO \cite{b6} (black arrows) and the proposed wflow-TartanVO (red arrows) architectures} \label{fig2}
\end{figure}

The proposed monocular learning-based VO system called wflow-TartanVO is built upon existing TartanVO \cite{b6} architecture, as illustrated in Fig. \ref{fig2}. The system receives two continuous frames ${I_t, I_{t+1}}$ as inputs, and outputs the estimated relative camera motion $\delta^{t+1}_t = (T, R)$, in which $T \in R^3$ and $R \in so(3)$ represents the 3D translation and 3D rotation respectively. Based on the theory of epipolar geometry, a VO system consists of two main steps. The first step is finding pixels correspondence across the frames, and the second step is computing the Essential matrix based on the correspondence, from which the camera motion is recovered. TartanVO \cite{b6} utilizes deep learning -based Matching Network $M_\theta$ to estimate the optical flow $F^{t+1}_t$ as correspondence. In the second step of recovering camera motion from correspondence, since TartanVO also attempts to solve the problem of generalization across different cameras, it incorporates camera intrinsic layers $K_c$ into correspondence result $F^{t+1}_t$. This creates a high-dimensional input, and thus another deep learning -based Pose Network $P_\phi$ is utilized, instead of using the Essential matrix.

The main novelty of the proposed learning-based VO is replacing optical flow result $F^{t+1}_t$, with a higher certainty optical flow $wF^{t+1}_t$ weighted by a transmission map estimated using deep learning models, as the input for the Pose module $P_\phi$. The normalization technique is also specified for the transmission map to adapt the amount of suppressing or emphasizing the motion, depending on the trajectory characteristics. 

\subsection{Deep learning-based estimation of Underwater Optical Imaging components}

Adopted from \cite{b11}, lightweight CNN modules are implemented to fit multiple components in the underwater optical image model. A-Net estimates the ambient light $A_c$ in Eq. \ref{eq:3} from the input image. The second stage T-Net estimates the inverse medium transmission map $T_{inv}$, which is $t^{-1}_c(x)$ in Eq. \ref{eq:3}. These underwater components estimation modules are trained separately from the main VO module, and thus, the trained A-Net and T-Net can be integrated into the existing pre-trained TartanVO model, without the need for fine-tuning.

\subsection{Attenuation-aware Weighted Optical Flow with Normalized Medium Transmission Map}
After estimation, the inverse transmission map $T_{inv}$ performs element-wise inverse to obtain the transmission map $T$. Then, it is normalized to the range $[0, 1]$, resembling the transmission quality percentage weight map for optical flow. Since the purpose is to suppress or emphasize the motion, the estimated medium transmission map is normalized into the new range $[1-\sigma, 1+\alpha-\sigma]$, with $\sigma$ calculated as:

\begin{equation} \label{eq:6}
\sigma = \frac{max(\alpha(1/T_{inv}))}{\beta}
\end{equation}

where $\alpha$, is a hyper-parameter controlling the spread of weight range, and $\beta$ controls the bias towards suppressing or emphasizing motion. To summarize, the normalized transmission map $T_{norm}$ is computed from the estimated $T_{inv}$:

\begin{equation} \label{eq:7}
T_{norm} = \alpha (1/T_{inv}) + 1 - \frac{max(\alpha(1/T_{inv}))}{\beta} 
\end{equation}

The final optical flow is obtained by performing the Hadamard product between the original optical flow and the normalized transmission map: $wF^{t+1}_t = F^{t+1}_t \odot T_{norm}$.

\begin{figure}[H]
\centering
\includegraphics[width=.7\linewidth]{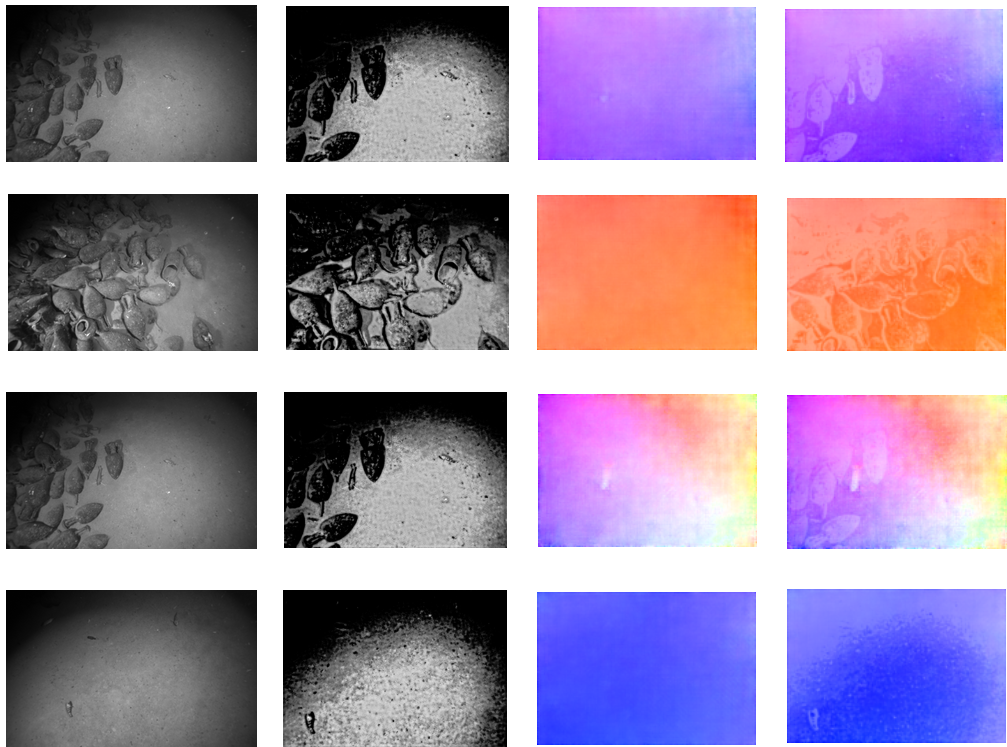}
\caption{Example results on 4 different underwater scenes (top to bottom) in Aqualoc \cite{b16}.} \label{fig4}
\end{figure}

The first row in Fig. \ref{fig4} shows an example of an underwater scene, and its normalized transmission map $T_{norm}$, the optical flow $F^{t+1}_t$ generated by PWC-Net \cite{b20}, and the proposed weighted optical flow $wF^{t+1}_t$. Although the original optical flow depicts motions of all pixels in a frame, which is true considering the camera movements for observing the scene, all motions seem to share approximately the same value and direction (same color and intensity), despite the camera's tilting angle, leading to different motion characteristics. For example, the regions closer to the camera (i.e.: nearby rocks, bottom image) should have higher motion, and vice versa, according to the motion parallax effect. Additionally, regions with higher degradation levels (i.e.: top corners) contain higher uncertainty, and thus should not share the same motion intensity as clearer regions. On the other hand, the proposed weighted optical flow considers these degraded regions to have less motion (more white value), and emphasize the motions of regions closer to camera (purple with higher intensity). Other scenes in motion examples are illustrated in the remaining rows of Fig. \ref{fig4}.

\section{Experiments} \label{section-5}

\begin{table*}[b]
\centering
\caption{Comparisons of trajectory length and number of poses between ORB-SLAM3 \cite{b3}, DSO \cite{b4}, TartanVO \cite{b6}, and the proposed wflow-TartanVO, with COLMAP reference.}
\label{tbl4}
\resizebox{.8\textwidth}{!}{%
\begin{tabular}{|l|rr|rr|rr|rr|rr|}
\hline
 & \multicolumn{2}{l|}{\textbf{COLMAP (ground truth)}} & \multicolumn{2}{l|}{\textbf{ORB-SLAM3}} & \multicolumn{2}{l|}{\textbf{DSO}} & \multicolumn{2}{l|}{\textbf{TartanVO}} & \multicolumn{2}{l|}{\textbf{wflow-TartanVO}} \\ \cline{2-11} 
\multirow{-2}{*}{\textbf{Scene}} & \multicolumn{1}{l|}{\textbf{Length (m)}} & \multicolumn{1}{l|}{\textbf{\# poses}} & \multicolumn{1}{l|}{\textbf{Length (m)}} & \multicolumn{1}{l|}{\textbf{\# poses}} & \multicolumn{1}{l|}{\textbf{Length (m)}} & \multicolumn{1}{l|}{\textbf{\# poses}} & \multicolumn{1}{l|}{\textbf{Length (m)}} & \multicolumn{1}{l|}{\textbf{\# poses}} & \multicolumn{1}{l|}{\textbf{Length (m)}} & \multicolumn{1}{l|}{\textbf{\# poses}} \\ \hline
\textbf{Aqualoc (sequence 6)} & \multicolumn{1}{r|}{16.212} & 1500 & \multicolumn{1}{r|}{1.426} & 56 & \multicolumn{1}{r|}{0.404} & 65 & \multicolumn{1}{r|}{{\ul 16.798}} & 1500 & \multicolumn{1}{r|}{{\color[HTML]{000000} \textbf{16.554}}} & 1500 \\ \hline
\textbf{Aqualoc (sequence 7)} & \multicolumn{1}{r|}{8.041} & 1500 & \multicolumn{1}{r|}{4.665} & 150 & \multicolumn{1}{r|}{6.663} & 165 & \multicolumn{1}{r|}{{\ul 8.154}} & 1500 & \multicolumn{1}{r|}{{\color[HTML]{000000} \textbf{8.065}}} & 1500 \\ \hline
\textbf{Aqualoc (sequence 8)} & \multicolumn{1}{r|}{35.785} & 1500 & \multicolumn{1}{r|}{2.993} & 80 & \multicolumn{1}{r|}{0.252} & 143 & \multicolumn{1}{r|}{{\ul 33.294}} & 1500 & \multicolumn{1}{r|}{{\color[HTML]{000000} \textbf{34.53}}} & 1500 \\ \hline
\textbf{Aqualoc (sequence 9)} & \multicolumn{1}{r|}{21.99} & 3000 & \multicolumn{1}{r|}{8.353} & 224 & \multicolumn{1}{r|}{7.35} & 226 & \multicolumn{1}{r|}{{\ul 19.714}} & 3000 & \multicolumn{1}{r|}{{\color[HTML]{000000} \textbf{21.516}}} & 3000 \\ \hline
\textbf{Aqualoc (sequence 10)} & \multicolumn{1}{r|}{19.02} & 1500 & \multicolumn{1}{r|}{4.002} & 101 & \multicolumn{1}{r|}{4.15} & 112 & \multicolumn{1}{r|}{{\ul 17.404}} & 1500 & \multicolumn{1}{r|}{{\color[HTML]{000000} \textbf{18.576}}} & 1500 \\ \hline
\textbf{SubPipe (chunk 3)} & \multicolumn{1}{r|}{372.785} & 16170 & \multicolumn{1}{l|}{N/A} & \multicolumn{1}{l|}{N/A} & \multicolumn{1}{l|}{N/A} & \multicolumn{1}{l|}{N/A} & \multicolumn{1}{r|}{{\ul 3282.656}} & 16170 & \multicolumn{1}{r|}{{\color[HTML]{000000} \textbf{1262.508}}} & 16170 \\ \hline
\end{tabular}%
}
\end{table*}

\begin{table*}[t]
\centering
\caption{Comparisons of ATE and RTE between ORB-SLAM3 \cite{b3}, DSO \cite{b4}, TartanVO \cite{b6}, and the proposed wflow-TartanVO, with COLMAPreference.}
\label{tbl5}
\resizebox{.8\textwidth}{!}{%
\begin{tabular}{|l|rr|rr|rr|rr|}
\hline
 & \multicolumn{2}{l|}{\textbf{ORB-SLAM3 (aligned)}} & \multicolumn{2}{l|}{\textbf{DSO (aligned)}} & \multicolumn{2}{l|}{\textbf{TartanVO}} & \multicolumn{2}{l|}{\textbf{wflow-TartanVO}} \\ \cline{2-9} 
\multirow{-2}{*}{\textbf{Scene}} & \multicolumn{1}{l|}{\textbf{ATE (m)}} & \multicolumn{1}{l|}{\textbf{RTE (m)}} & \multicolumn{1}{l|}{\textbf{ATE (m)}} & \multicolumn{1}{l|}{\textbf{RTE (m)}} & \multicolumn{1}{l|}{\textbf{ATE (m)}} & \multicolumn{1}{l|}{\textbf{RTE (m)}} & \multicolumn{1}{l|}{\textbf{ATE (m)}} & \multicolumn{1}{l|}{\textbf{RTE (m)}} \\ \hline
\textbf{Aqualoc (sequence 6)} & \multicolumn{1}{r|}{0.2831} & 0.0464 & \multicolumn{1}{r|}{1.5336} & 0.0992 & \multicolumn{1}{r|}{{\color[HTML]{000000} {\ul 0.2788}}} & {\color[HTML]{000000} {\ul 0.0212}} & \multicolumn{1}{r|}{{\color[HTML]{000000} \textbf{0.2431}}} & {\color[HTML]{000000} \textbf{0.021}} \\ \hline
\textbf{Aqualoc (sequence 7)} & \multicolumn{1}{r|}{0.602} & 0.0261 & \multicolumn{1}{r|}{0.2022} & {\color[HTML]{000000} \textbf{0.0078}} & \multicolumn{1}{r|}{{\color[HTML]{000000} {\ul 0.2161}}} & 0.0081 & \multicolumn{1}{r|}{{\color[HTML]{000000} \textbf{0.1909}}} & {\color[HTML]{000000} {\ul 0.008}} \\ \hline
\textbf{Aqualoc (sequence 8)} & \multicolumn{1}{r|}{3.9508} & 0.4464 & \multicolumn{1}{r|}{3.3708} & 0.263 & \multicolumn{1}{r|}{{\color[HTML]{000000} {\ul 1.3655}}} & {\color[HTML]{000000} \textbf{0.0375}} & \multicolumn{1}{r|}{{\color[HTML]{000000} \textbf{1.2605}}} & {\color[HTML]{000000} {\ul 0.038}} \\ \hline
\textbf{Aqualoc (sequence 9)} & \multicolumn{1}{r|}{2.446} & 0.077 & \multicolumn{1}{r|}{2.4448} & 0.063 & \multicolumn{1}{r|}{{\color[HTML]{000000} {\ul 0.7877}}} & {\color[HTML]{000000} \textbf{0.0106}} & \multicolumn{1}{r|}{{\color[HTML]{000000} \textbf{0.6669}}} & {\color[HTML]{000000} {\ul 0.0111}} \\ \hline
\textbf{Aqualoc (sequence 10)} & \multicolumn{1}{r|}{3.1412} & 0.1967 & \multicolumn{1}{r|}{2.7016} & 0.1276 & \multicolumn{1}{r|}{{\color[HTML]{000000} {\ul 0.4096}}} & {\color[HTML]{000000} \textbf{0.0176}} & \multicolumn{1}{r|}{{\color[HTML]{000000} \textbf{0.3617}}} & {\color[HTML]{000000} {\ul 0.0185}} \\ \hline
\textbf{SubPipe (chunk 3)} & \multicolumn{1}{l|}{N/A} & \multicolumn{1}{l|}{N/A} & \multicolumn{1}{l|}{N/A} & \multicolumn{1}{l|}{N/A} & \multicolumn{1}{r|}{{\color[HTML]{000000} {\ul 45.7378}}} & {\color[HTML]{000000} {\ul 0.2427}} & \multicolumn{1}{r|}{{\color[HTML]{000000} \textbf{36.1032}}} & {\color[HTML]{000000} \textbf{0.0991}} \\ \hline
\end{tabular}%
}
\end{table*}

This section compares camera poses estimation of proposed attenuation-aware learning-based wflow-TartanVO with other SLAM baselines such as geometry-based ORB-SLAM3 \cite{b3}, DSO \cite{b4}, and learning-based TartanVO \cite{b6}, on Aqualoc \cite{b16} (sequence 6, 7, 8, 9, 10) and SubPipe \cite{b17} (chunk 3) datasets.

ORB-SLAM3 is an indirect VO approach using a sparse set of ORB descriptors to be tracked across frames under the condition of sufficient texture, along with loop closure to correct scale drifting. DSO is a direct VO tracking pixel intensity gradients across frames, thus using more pixel information instead of a sparse set of key points, under the condition of sufficient overlap between frames. Meanwhile, the learning-based TartanVO uses optical flow to track all pixels across a frame, but it depends on the accuracy of optical flow estimation, whose partial motion values might be uncertain due to underwater degradation factors. Thus, wflow-TartanVO is proposed to provide a weighted optical flow with higher certainty for TartanVO. The hyper-parameters ($\alpha$, $\beta$) controlling spread and bias, as in Eq. \ref{eq:7}, were set for Aqualoc sequence 6, 7, 8, 9, 10 and SubPipe chunk 3 as follows: $(0.1,4), (0.25,4), (0.25,4), (0.25,4), (0.25,4), (1.0,4)$. The Aqualoc sequence 6 has a relatively straight trajectory compared to others, thus a smaller weight spread was found to provide better results compared to higher values.

\begin{figure}[t]
     \centering
     \begin{subfigure}[b]{0.24\textwidth}
         \centering
         \includegraphics[width=\textwidth]{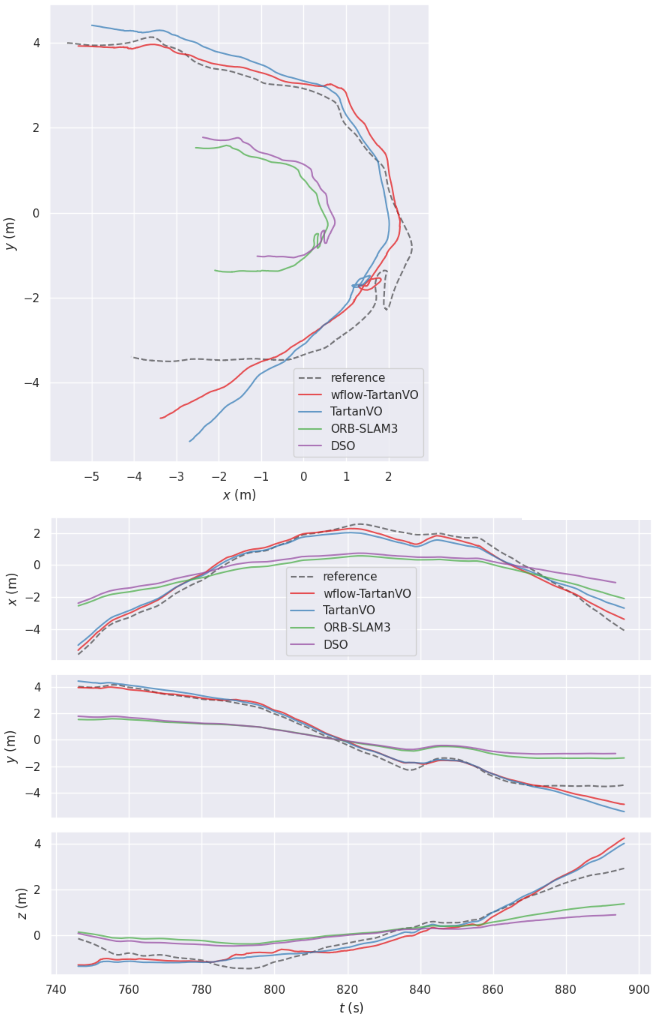}
         \caption{Aqualoc - Sequence 9}
         \label{fig7-1}
     \end{subfigure}
     \hfill
     \begin{subfigure}[b]{0.24\textwidth}
         \centering
         \includegraphics[width=\textwidth]{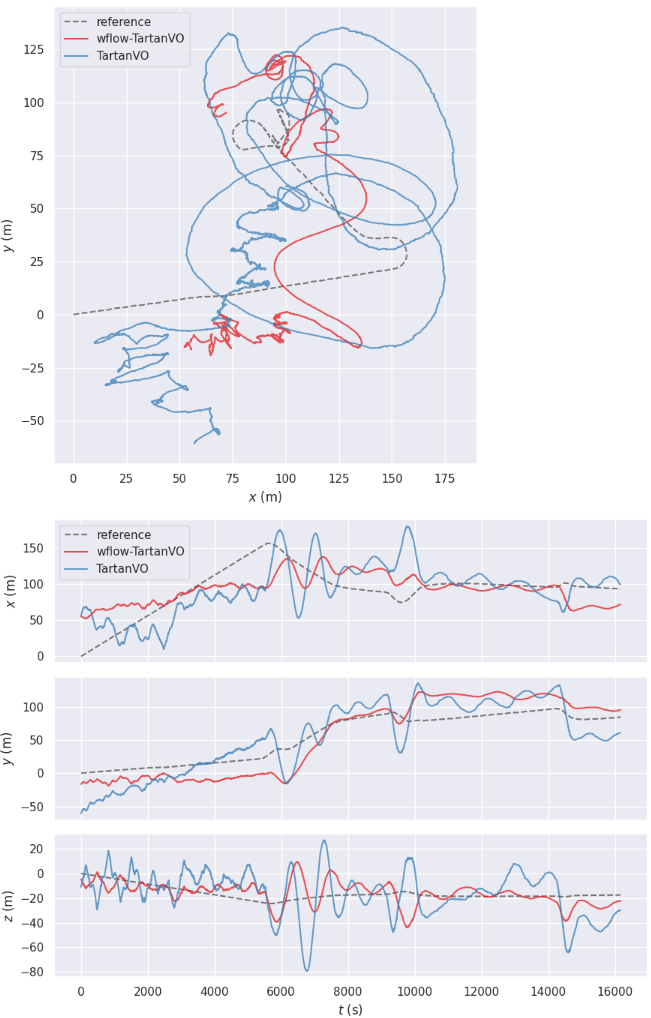}
         \caption{SubPipe - Chunk 3}
         \label{fig7-3}
     \end{subfigure}
    \caption{Visualization of trajectories on x-y plane (top) and x,y,z axes (bottom) of Aqualoc \cite{b16} and RGB SubPipe \cite{b17}}
    \label{fig7}
\end{figure}

Figure \ref{fig7} shows example trajectory results on Aqualoc sequence 9 and SubPipe chunk 3. It indicates that both geometry-based methods suffered from major scale-drift on Aqualoc sequences, which can be explained by the low number of estimated poses (or number of key frames) as shown in Table \ref{tbl4}. Low number of key frames led to inadequate loop closure detection to correct scale drifting for ORB-SLAM3, and also limited scene coverage or lack of redundancy to provide sufficient frames overlap for DSO. Meanwhile on SubPipe dataset, both geometry-based methods failed tracking due to challenging imaging conditions with high texture uniformity, leading to a lack of features and gradients. Since learning-based VO can utilize all pixels and leverage from higher-level representations of image, TartanVO and the proposed wflow-TartanVO could obtain fairly accurate trajectories in both scale and shape for Aqualoc sequences, and were also able to estimate all poses in challenging Subpipe chunk 3. The trajectories estimated by wflow-TartanVO could align better with the ground truth due to the attenuation-aware weighted optical flow and appropriate hyper-parameters tuning.

The quantitative comparisons for all sequences of the two datasets can be found in Table \ref{tbl4} and Table \ref{tbl5}. The best and second best results are highlighted in bold and underlined. Table \ref{tbl4} illustrates the estimated trajectory length with the number of poses (key frames), and shows that learning-based VO methods estimate trajectory length much closer to ground truth due to minor scale drift, compared to geometry-based VO. Table \ref{tbl5} shows Absolute Trajectory Error (ATE) and Relative Trajectory Error (RTE) comparisons. ATE computes the root mean squared error between the whole estimated trajectory and its reference, whereas RTE selects only a sub-trajectory, computes its RMSE, repeats for other sub-trajectories, and then finally collects the errors. The table reveals that the proposed wflow-TartanVO outperformed other baselines in all datasets for ATE metric, but was surpassed by TartanVO for RTE metric in some trajectories. This might be due to TartanVO generally had smoother trajectory, but underfitted reference trajectory shape, as seen from Aqualoc sequence 9 in Fig. \ref{fig7}. Smoother sub-trajectories led to lower overall RTE in some sequences. On the other hand, wflow-TartanVO had more varying trajectories, but corresponded better to the reference trajectory. However, it sometimes resulted in more intense variations, due to weighted optical flow with manual tuning hyper-parameters. This should be addressed in our future work.

\section{Conclusion} \label{section-5}
In conclusion, this paper presented wflow-TartanVO, a novel approach to improve learning-based monocular visual odometry (VO) for AUVs in underwater environments. By integrating principles of underwater optical imaging, optical flow estimation was manipulated using a normalized medium transmission maps, addressing the challenge of motion uncertainty posed by underwater light scattering and absorption. Experimental results demonstrated the improved performance of wflow-TartanVO over existing VO methods, as evidenced by considerably reduced ATE across different real-world underwater datasets. The adaptability of light-weight wflow-TartanVO, requiring no fine-tuning of pre-trained VO model, underscored its versatility and potential for real-world deployment. Future work includes optimizing current hyper-parameters from manual tuning to be learnable based on trajectory variations, and applying underwater image enhancement preprocessing step.

\end{document}